\documentclass[10pt,journal,compsoc]{IEEEtran}

\usepackage{cite}
\usepackage{amsmath,amssymb,amsfonts}
\usepackage{algorithmic}
\usepackage{graphicx}
\usepackage{textcomp}
\usepackage{multirow}
\usepackage{xcolor}
\usepackage{subfigure}
\usepackage{caption}
\usepackage{float}
\usepackage{breqn}

\if CLASSOPTIONcompsoc
  \usepackage[nocompress]{cite}
  \usepackage{cite}
\fi
\ifCLASSINFOpdf
\else
\fi
\begin{document}
%
\title{EXPRESSNET: An Explainable Residual Slim Network for Fingerprint Presentation Attack Detection}
%
%
%
%

\author{Anuj Rai,~\IEEEmembership{}
        Somnath Dey,~\IEEEmembership{Senior Member IEEE}
    
\IEEEcompsocitemizethanks{\IEEEcompsocthanksitem Anuj Rai and Somnath Dey are with the Department of Computer Science and Engineering, Indian Institute of Technology Indore, Indore, India, 453552.\protect\\
E-mail: phd1901201003@iiti.ac.in, somnathd@iiti.ac.in
\IEEEcompsocthanksitem}
\thanks{}}

%
%

\markboth{}%
{Shell \MakeLowercase{\textit{et al.}}: }
%

\IEEEtitleabstractindextext{%
\begin{abstract}
Presentation attack is a challenging issue that persists in the security of automatic fingerprint recognition systems. This paper proposes a novel explainable residual slim network that detects the presentation attack by representing the visual features in the input fingerprint sample. The encoder-decoder of this network along with the channel attention block converts the input sample into its heatmap representation while the modified residual convolutional neural network classifier discriminates between live and spoof fingerprints. The entire architecture of the heatmap generator block and modified ResNet classifier works together in an end-to-end manner. The performance of the proposed model is validated on benchmark liveness detection competition databases i.e. Livdet 2011, 2013, 2015, 2017, and 2019 and the classification accuracy of 96.86\%, 99.84\%, 96.45\%, 96.07\%, 96.27\% are achieved on them, respectively. The performance of the proposed model is compared with the state-of-the-art techniques, and the proposed method outperforms state-of-the-art methods in benchmark protocols of presentation attack detection in terms of classification accuracy.
\end{abstract}

\begin{IEEEkeywords}
Fingerprint Biometrics, Explainable Deep Learning, Presentation Attack Detection.
\end{IEEEkeywords}}
\maketitle

\IEEEdisplaynontitleabstractindextext

\IEEEpeerreviewmaketitle

\IEEEraisesectionheading{\section{Introduction}\label{sec:introduction}}
\IEEEPARstart{A}{n} Automatic Fingerprint Recognition System (AFRS) is a user-friendly and cost-effective solution for biometric-based person recognition. It takes less time, computing resources and human effort to verify a person than other biometric recognition systems. Due to its ease of use and automation, AFRS is being used for verification or authentication of a person in security-related applications such as Aadhar verification, airports \cite{reference1}, international borders, etc. Its usage in such security-sensitive applications makes it vulnerable to various threats. A Presentation Attack (PA) is one of them which is imposed by creating an artifact of a genuine user's finger and presenting it to the sensing device of an AFRS. The PAs can be created in two ways i.e. non-cooperative method of spoofing and cooperative method of spoofing. In the non-cooperative method, the latent fingerprint left on a surface is captured and then fabricated using spoofing material after digitization. On the other side,  the user itself provides an impression of their fingers to create the spoof in the cooperative method. Apart from this, the discovery of novel spoofing materials also imposes a big challenge to the security of AFRS. These materials are used to fabricate more realistic artifacts of fingers. Fingerprint Presentation Attack Detection (FPAD) is a countermeasure to PAs. The FPAD methods can be classified into two broad categories that are hardware-based methods and software-based methods. Hardware-based methods require additional devices for the measurement of the natural properties of the finger such as temperature, pulse rate and humidity which makes them costly. On the other hand, software-based methods require only the fingerprint sample which makes them user-friendly and cost-effective. Therefore, our focus is on the development of a software-based method that will be able to detect the PAs created with the help of known as well as unknown spoofing materials.

The state-of-the-art software-based methods are further classified as perspiration and pore-based methods \cite{b7_choi1, b25_marasco, b11}, statistical and handcrafted features-based methods \cite{xia_2,b13_xia1,b_Gragnaniello,b37_deepika,b5_sharma1,b1_dubey,b23_kim,b19_abhyankar} and deep learning-based methods \cite{b10_anusha,b26_arora,b14_chugh1,b15_chugh2,b4_uliyan, Nogueira,b47_jung2,Spinoulas}. Perspiration-based methods are proven to be insufficient because this property is affected by external temperature and other environmental factors. Along with this limitation, the feature extraction process of these methods requires multiple impressions of the same finger which makes it less user-friendly. Pore-based methods require the input samples to be of high-resolution ($>$1000 pixels per inch) which increases the cost of the FPAD system. Similarly, the quality of the sensing device impacts the performance of the statistical and handcrafted feature-based methods. In recent times, deep learning approaches have been adopted by various researchers due to their superior image classification capability. A set of convolutional filters possessed by them extracts minute features from input fingerprint samples. However, Convolutional Neural Networks (CNN) have the unmatched capability of extracting the discriminating features but they do not exhibit the same capability on fingerprint databases. The lack of texture and color information in fingerprint images is one of the possible reasons behind this. The depth of these networks makes them suffer from the vanishing gradient due to the lack of discriminating information. Hence some pre-processing is required in fingerprint databases to get good classification results.  \\
In this paper, we propose a novel end-to-end architecture that consists of a heatmap generator and a modified ResNet classifier. The Heatmap generator is composed of an encoder-decoder block and a channel attention block. It converts the input sample into a heatmap by emphasizing the important features present in an input fingerprint sample.
The encoder-decoder block highlights the features present in the region of interest in an image while the channel attention block finds discriminant features in the sample. The outcome of these aforementioned blocks is a single-channel heatmap which is fed to the modified ResNet classifier for the classification. The ResNet architecture \cite{resnet} is modified to make it less computationally expensive while being trained and tested on the fingerprint samples. The modification is done by removing the redundant convolutional blocks while maintaining their spatial properties and reducing the number of learnable parameters as well. The proposed EXPlainable RESidual Slim NETwork (EXPRESSNET) model is validated using Liveness Detection Competition (LivDet) 2011, 2013, 2015, 2017 and 2019 databases. It outperforms existing FPAD methods in intra-sensor same-material and unknown-material protocols. The main contributions of this paper are discussed as follows. \\

1. To the best of our knowledge, we are the first to introduce the concept of explainability of deep CNN in the area of FPAD.

2. The proposed model highlights the driving features of input fingerprint samples by converting them into a single-channel heatmap. In this way, discriminating features such as wetness, ridge and valley clarity and scars are highlighted for better classification.

3. The proposed heatmap generator block can be attached to any CNN classifier to enhance its classification performance.

4. The spatial properties of ResNet's feature maps are preserved along with a reduction in the number of learnable parameters by proposing modifications in the original ResNet architecture.

5. A detailed comparison of the proposed model has been done against the spoofs created using cooperative and non-co-operative subjects as well as known and unknown spoofing materials.

The remainder of this paper is organized as follows. Section \ref{related work} discusses existing methodologies suggested by various researchers. Section \ref{proposed work} describes the design and working of the proposed architecture. In section \ref{experimental results}, experimental results, as well as comparative analysis are given. Finally, the paper is concluded in section \ref{conclusion}.

\section{Related Work} \label{related work}
FPAD is an essential tool for the AFRS to deal with PAs. As a countermeasure to PAs, researchers have proposed a variety of software-based solutions, which may be further categorized as pore and perspiration-based methods, statistical and handcrafted feature-based methods and deep learning-based methods. This section discusses the most recent approaches that fall into these categories, as well as their advantages and limitations.
\subsection{\textbf{Perspiration and pore based-methods}}
The presence of small holes or pores in human skin causes perspiration in fingers. This natural property is not present in the spoofs fabricated with different materials. An initial study was proposed by Derakshani et al. \cite{b27_derakshini}. They utilized the diffusion pattern of sweat as a feature to discriminate between live and spoof fingerprints. Later, Abhyankar et al. \cite{b19_abhyankar} proposed a wavelet-based method that utilizes the sweat feature of the fingerprint to detect PAs. Since, the pores are hard to reflect in the spoofs at the time of fabrication, the number of pores may differ in a live fingerprint and its spoofs created with different materials. This dissimilarity is utilized as a discriminating feature by Espinoza \cite{b16_espinoza}. The proposed method is validated using a custom-made fingerprint database. Similarly, Marcialis et al. \cite{b17_marcialis} captured two fingerprint impressions at an interval of five-second and then detects the pores in both impressions. The proposed method utilizes the number of pores present in both impressions as a feature for detecting PAs. The proposed method is validated using a custom-made fingerprint database that consists of 8960 live and 7760 spoof images.
Though, the perspiration pattern is used for the detection of PAs, its presence depends on the atmosphere temperature. A live finger in a dry environment does not exhibit this property which causes the discard of the live sample by the FPAD system working on this feature. Moreover, the extraction of pores has been shown to be expensive since the fingerprint sensor must be capable of capturing high-definition samples ($>=$1000 pixels per inch). For the reasons stated above, perspiration and pore-based approaches are less user-friendly and cost-effective.
\subsection{\textbf{Statistical and handcrafted feature based-methods}}
The skin of a finger and its counterpart, the fabricated spoofs, have different natural properties such as color, wetness and elasticity level which are reflected in the quality of the samples captured with fingerprint sensors. Statistical and handcrafted feature-based methods use quality features of the fingerprints for the detection of their liveness. 
Choi et al. \cite{b7_choi1} extracted histogram, directional contrast, ridge thickness and ridge signal features for detecting PAs. They utilized these features for the training of an SVM classifier. The proposed method is validated using the custom-made fingerprint database. Similarly, Park et al. \cite{b41_park2} utilized statistical features including standard deviation, variance, skewness, kurtosis, hyper-skewness and hyper-flatness along with three additional features i.e. average brightness, standard deviation and differential image for the training of SVM to detect PAs. They validated their method using the ATVSFFp database which contains 272 real fingerprints and 270 spoof fingerprint samples.
Further, Xia et al. \cite{b13_xia1} extracted second and third-order occurrence of gradients from fingerprint samples. They used these features for the training of the SVM classifier. The proposed method is validated using LivDet 2009 and 2011 databases. In another work \cite{xia_2}, Xia et al. suggested a novel image descriptor that extracts intensity variance along with gradient properties of the fingerprint samples to form a feature vector. This feature vector is further used for the training of the SVM classifier. The proposed work is validated using LivDet 2011 and 2013 databases.
Yuan et al. \cite{b44_yuan3} in continuation to the work of \cite{xia_2}, proposed a method that utilizes  gradient property for the detection of the PAs. It creates two co-occurrence matrices using the Laplacian operator that compute image gradient values for different quantization operators. Further, The matrices are utilized as a feature vector for the training of the back-propagation neural network. The suggested method is validated using LivDet 2013 database. Since the live finger and its spoof have different levels of elasticity, it is reflected in the varying width of the ridges and valleys as well as the quality of their image samples. Sharma et al. \cite{b5_sharma1} extracted some quality features such as Ridge and Valley Clarity (RVC), Ridge and Valley Smoothness (RVS), Frequency Domain Analysis (FDA) and Orientation Certainty Level (OCL) which are combined together for the training of Random-Forest classifier. The proposed method is validated on LivDet 2009, 2011, 2013 and 2015 databases.
Sharma et al. \cite{b37_deepika} suggested a novel feature named the Local Adaptive Binary Pattern (LABP) which is the modification to the existing Local Binary Pattern (LBP). They have combined this feature with existing BSIF and Complete Local Binary Pattern (CLBP) and used them for the training of the SVM classifier. The proposed method is validated on LivDet 2009, 2011, 2013 and 2015 databases. Ghiani et al.\cite{b12_ghiani} utilized BSIF which is obtained by applying a set of pre-defined filters whose output is then converted to a binary sequence. This binary sequence is used as a feature vector for the training of the SVM classifier. The proposed method is tested on LivDet 2011 database. 
The varying elasticity of the live fingers and corresponding spoofs causes a significant difference in their shapes and textures also. Further, Dubey et al. \cite{b1_dubey} suggested a shape and texture feature-based method. They utilized Speeded Up Robust Feature (SURF) and Pyramid extension of Histogram of Gradient (PHOG) to extract shape information from the fingerprint sample. Along with the aforementioned features, the Gabor wavelet is used by them to extract the texture information. The proposed method is validated using LivDet 2011 and 2013 databases.
Ajita et al. \cite{b2_rattani1} proposed a novel method for the detection of PAs created with unknown materials. They suggested the use of an Adaptive-Boost (AdaBoost) multi-class classifier that classifies an input fingerprint as live, spoof and unknown. The Fingerprint samples detected as `unknown' are further used to train the classifier to detect their presence in the future. The proposed method is tested on LivDet 2011 database.
In continuation to their previous work \cite{b2_rattani1}, Ajita et al. \cite{b3_rattani2} suggested the use of a Weibull-calibrated SVM classifier for the detection of PAs. This SVM is a combination of 1-class as well as binary SVM. This modification shows a significant improvement as compared with the results on LivDet 2011 database.
Kim et al. \cite{b23_kim} proposed a novel image descriptor that utilizes the local coherence of the fingerprint sample as a feature for the training of SVM. The proposed method is validated using ATVSFFp and LivDet 2009, 2011, 2013 and 2015 databases.\\
The efficacy of these methods depends on the quality of the input fingerprint sample which further depends on the sensing device. Some of the aforementioned methods \cite{b5_sharma1, b37_deepika, b23_kim} have shown remarkable performance against the PAs created using known fabrication materials but do not resemble the same against the spoofs created using novel materials.
\vspace{-2.0 mm}
\subsection{\textbf{Deep learning based-methods}} Deep CNNs can extract minute information from image samples since they have convolutional layers. These models have shown excellent classification capabilities when evaluated on imagenet \cite{imagenet}, CIFAR \cite{cifar} and MNIST \cite{mnist} databases. This benefit led researchers to use CNNs in the detection of PAs as well. This section discusses state-of-the-art deep learning-based FPAD methods.  
Arora et al. \cite{b26_arora} proposed a robust framework to detect presentation attacks in fingerprint biometric systems that involves contrast enhancement using histogram equalization. Fingerprint samples after pre-processing are fed to the VGG classifier. The proposed work is validated on benchmark fingerprint databases which include FVC 2006, ATVSFFp, Finger vein data-set, LivDet 2013 and 2015 databases. Similarly, Nogueira et al. \cite{Nogueira} utilized pre-trained CNN architectures using transfer learning. Their method involves existing deep CNN architectures such as VGG, Alexnet and CNN with SVM. The proposed method is tested on LivDet 2009, 2011 and 2013 databases.
Uliyan et al.\cite{b4_uliyan} proposed deep features-based methods for the detection of PAs. It utilizes a Deep Boltzmann Machine (DBM) for the extraction of features from fingerprint images. DBM has been utilized by them to find the complex relationship among the features. The proposed work is validated using benchmark fingerprint databases.
Chugh et al. \cite{b14_chugh1} suggested a deep learning-based method that uses minutiae-centered fingerprint patches for the training and testing of a MobileNet classifier. A fingerprint is divided into a finite number of patches based on the number of minutiae points present in it. Extracted patches are fed to a CNN model which generates a liveness score for every patch. The liveness score for an input sample is computed using score-level fusion. This proposed method is validated using LivDet 2011, 2013 and 2015 databases and Michigan State University's (MSU) FPAD database.
Since, novel fabrication materials are discovered every day, it is hard to generalize an FPAD model to perform FPAD in an open-set or unknown-material protocol.
In continuation of their previous work \cite{b15_chugh2}, Chugh et al. \cite{b14_chugh1} suggested another method for the detection of spoofs fabricated using unknown materials. They proposed an image synthesis technique to create new fingerprint patches which contribute to better training of the MobileNet classifier. The proposed method is validated using LivDet 2017, ATVSFFp and MSU-FPAD databases. Zhang et al. \cite{b46_zhang2} suggested a CNN architecture that outperforms all the feature-based methods in terms of classification accuracy. They proposed an architecture that consists of a series of improved residual connected blocks. This modified architecture results in the detection of PAs without over-fitting and less computing time. The proposed method is validated on Livdet 2013 and 2015 databases.
\begin{figure*}[t]
	\centering
	\resizebox{1.30\textwidth}{!}{
		{
			
			\includegraphics[]{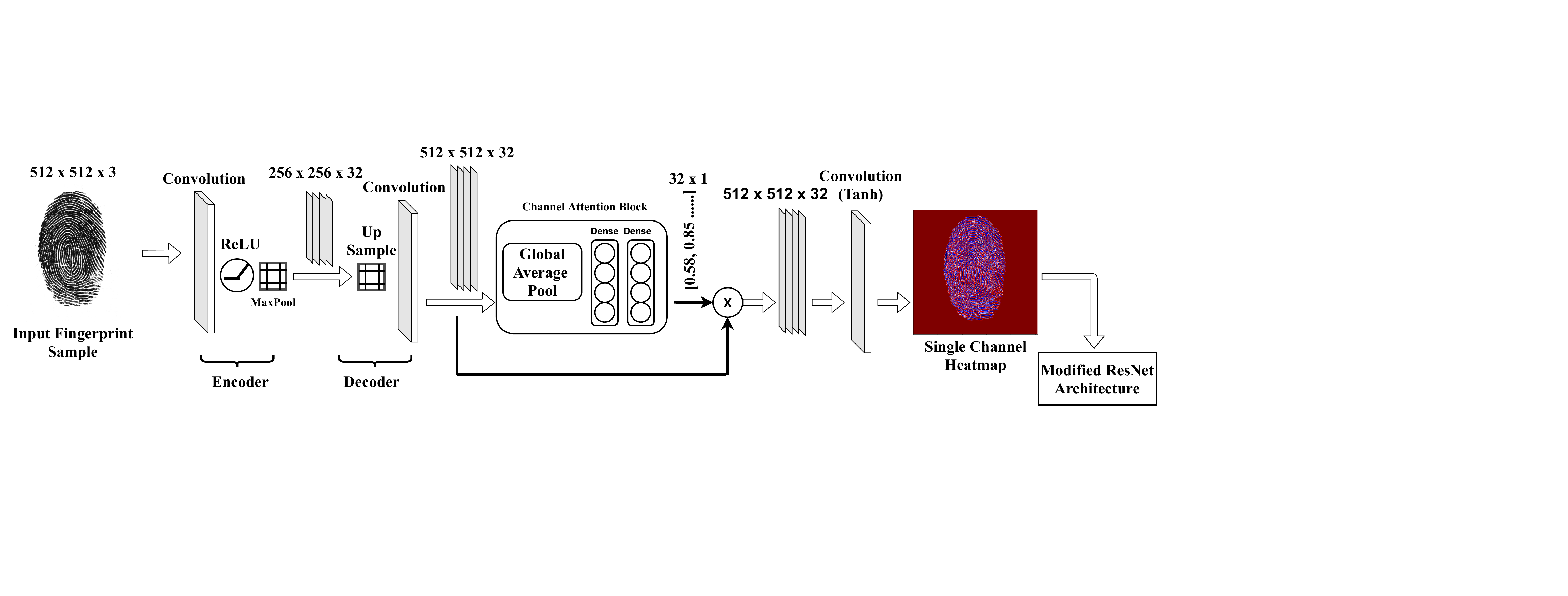}
	}}
 \vspace{-25mm}
	\caption{Block diagram of EXPRESSNET architecture}
	\label{EXPRESSNET_1}
 \vspace{-1mm}	
\end{figure*}
\subsection{Explainability in Deep Learning} The term explainability refers to any information that helps the user understand the pattern of the decisions made by the deep learning model for the input samples belonging to different classes. In recent times, various surveys \cite{EXP2}, \cite{EXP3} have been proposed to enlighten this area. The Explainability in DNNs can be achieved in three ways including Visualization methods, Model distillation and Intrinsic-methods. Visualization methods, being applied to the image classifiers, are further classified as backpropagation-based methods \cite{EXP5}, activation maximization methods \cite{EXP6}, deconvolution methods \cite{EXP4} and layer-wise relevance propagation-based methods \cite{EXP7}, etc. Deconvolution methods utilize inverse convolution operations to visualize high-layer features present in the input image samples. Amir et al. \cite{EXP1} utilized the deconvolution method in an attempt to emphasize the important features present in the input sample. The proposed method is tested on CIFAR, MNIST and tiny-imagenet databases. The performance is compared with state-of-the-art explainability methods. This method performs well on images belonging to different classes based on the shape, color and texture of the objects present in them. Since live and spoof fingerprint samples can not be discriminated based on these features the deconvolution method is required to be enhanced for fingerprint databases.

The detailed literature review concludes that the deep learning-based methods have shown remarkable performance while being applied in the area of image classification problems but they are not sufficient while being utilized for live and spoof fingerprint samples. One of the possible reasons may be the limited amount of discriminating features in fingerprint samples. We have developed a novel approach that highlights the key features that play a vital role in the discrimination of live and spoof fingerprint samples without imposing computational overhead on the entire FPAD system which is discussed in the following sections.

\section{Proposed Work \label{proposed work}}
In this paper, we propose a novel architecture to detect PAs by generating heatmaps. The architecture, shown in Fig. \ref{EXPRESSNET_1} consists of the encoder-decoder and the channel attention block for heatmap generation and modified ResNet for classification. The first component highlights the regions as well as discriminating features that play a vital role in the classification process. In this way, the classifier is empowered for better classification of the input samples. The details of the components of the EXPRESSNET architecture are mentioned in the following subsections.

			

\begin{figure*}[ht]

	\centering
	
	\resizebox{0.9\textwidth}{!}{
		{
			
			\includegraphics[]{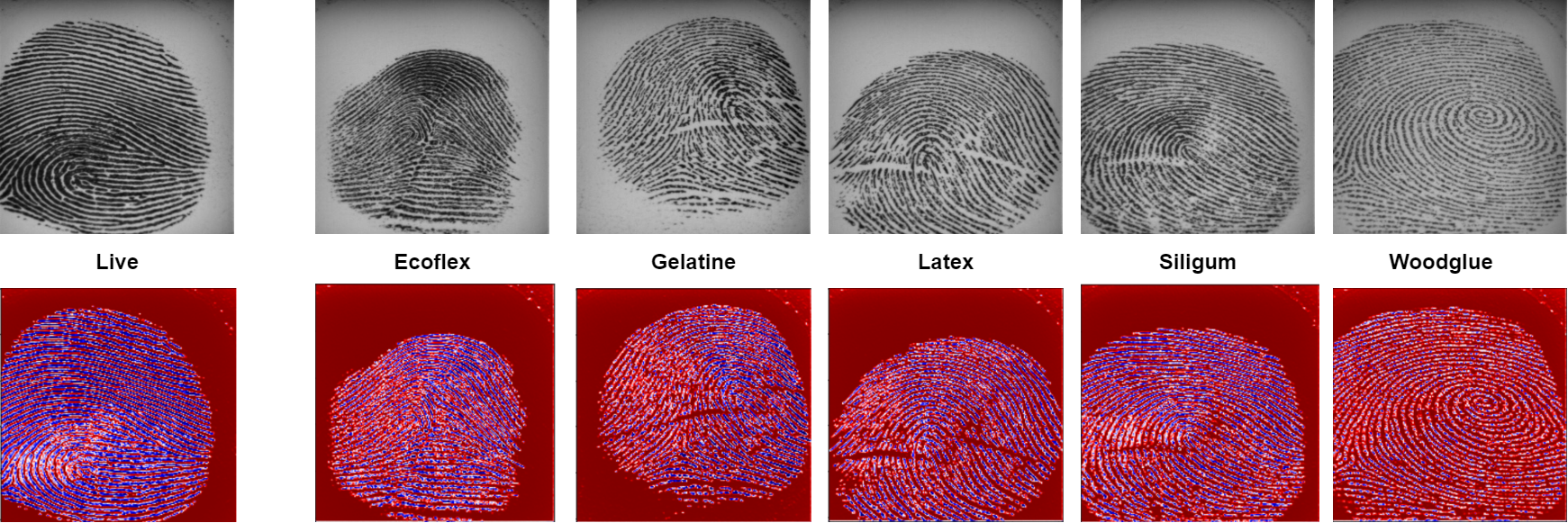}
	}}
	\caption{Live and spoofs fabricated with various materials along with their generated heatmaps by the proposed heatmap generator}
	\label{Heatmaps}
 \vspace{-1mm}	
\end{figure*}

\subsection{Preprocessing} The sample captured with different sensing devices has different spatial dimensions. To overcome this problem, fingerprint samples are resized into the size of $512 \times 512$. This modification, in turn, increases model training time while having no effect on the number of trainable parameters.
\subsection{Heatmap Generator Block}
The resized input sample is passed to the heatmap generator block which constitutes of encoder-decoder, channel attention block and heatmap generation layers. The details of the aforementioned blocks are given in the following subsections.
\subsubsection{\textbf{Encoder-Decoder Block}} The proposed encoder-decoder block first down-size and then up-sizes the input feature maps to highlight the features present in them. In other words, the encoder extracts relevant information and the decoder shows the driving features present in the feature maps while retaining their spatial properties. The encoder part is composed of convolutional operation along with pooling operation. The convolutional filter extracts feature from the sample while the poling operation downsamples the input sample. The output of the encoder block is formulated as Eq. (\ref{encoder}).
\begin{equation}
\label{encoder}
Encoder_{out}  = Maxpool\bigg[\sum_{X,x=0}^{M,m}\sum_{Y,y=0}^{N,n}I_{X,Y} \times K_{x,y}\bigg]
\end{equation}
Here, $I_{X, Y}$ denotes the input fingerprint sample of dimension $M \times N$ and $K_{x,y}$ denotes the convolutional filter with size $x \times y$. After convolution, the max-pooling operation is used to downsample the output feature maps. $Encoder_{out_f}$ denote the output feature maps. 
The output of the encoder is passed to the decoder to enhance the features. In \cite{EXP1} the decoder consists of transposed convolution operator which is higher in terms of computational cost. To keep this cost low, we have constituted the decoder block using an up-sample operation followed by the convolutional operation. The decoder block can be formulated as Eq. (\ref{decoder}).
\begin{dmath}
\label{decoder}
Decoder_{out}  = \bigg[\sum_{X,x=0}^{M,m}\sum_{Y,y=0}^{N,n}\big(\big(Upsample(Encoder_{out}\big) \times K_{x,y}\big)\bigg]
\end{dmath}
Here, $Decoder_{out}$ is the output of the encoder-decoder block which is a set of `$f$' feature maps of size $M \times N$ each. In this model, the value of `$f$' is kept as 32. These output feature maps have highlighted pixels that contribute to the classification of the input sample. The feature maps are fed to the channel attention block which is described in the following subsection.


\subsubsection{\textbf{Channel Attention Block (CAB)} \label{CAB}} The CAB produces an attention map that exploits the inter-channel relationship of features. The goal of the encoder-decoder block is to find ``where" the important feature is present while the CAB is responsible for finding ``what" is important in the image.
The calculation of channel attention is formulated as per Eq. (\ref{expresseq3}).
\begin{equation}
\label{expresseq3}
CAB_{out}  =   MLP(AveragePool(Decoder_{out}))
\end{equation}
Here, Multi-Layer Perceptron (MLP) is a collection of two dense layers. The formation of MLP is denoted with Eq. (\ref{expresseq4}).
\begin{equation}
\label{expresseq4}
MLP  =  ReLU(W_{1}\sigma(W_{0}()))
\end{equation}
Here, $W1$ and $W0$ represent the weights of fully-connected layers and ReLU and Sigmoid are the activation functions applied to those layers respectively.
The channel attention map is then multiplied by the feature maps generated by the encoder-decoder block. The feature maps with highlighted information are then merged together to form a single-channel heatmap. A convolutional filter is utilized for the same which is mentioned in the following subsection.
\subsubsection{\textbf{Heatmap Generation Layer}}The output of the channel attention block is a set of feature maps that have important features highlighted. These feature maps are further to be merged to form a single-channel heatmap. For the same, a convolutional filter is used that takes `f' feature maps as input and produces a single heatmap as an output. The formulation of the same is given as Eq. \ref{heatmap}. This operation is followed by the Tanh activation function that maps input values in the range (-1 to +1). Figure \ref{Heatmaps} depicts live and spoof fingerprint samples belonging to LivDet 2011, biometrika dataset and respective heatmaps generated by the heatmap generator module. 
\begin{equation}
Heatmap = Tanh\bigg[\sum_{X,x=0}^{M,m}\sum_{Y,y=0}^{N,n}(Decoder_{out_f}\times CAB_{out})\bigg]
\label{heatmap}
\end{equation}

As seen in Fig. \ref{Heatmaps}, it is evident that the discriminating features such as wetness, noise, scar, clarity of ridges and valley widths are highlighted by the proposed heatmap generator. The output heatmap is fed as an input to the classifier. For the classification of fingerprint heatmaps, Residual CNN is opted as a classifier and to reduce the computational cost, its architecture has been modified. The details of the original and modified ResNet classifiers are mentioned in the following subsection.
\subsection{\textbf{Modified Residual CNN (Slim-ResNet) Classifier}} 
\begin{figure}[!t]
	\centering
	
	\resizebox{0.31\textheight}{!}{
		{
			
			\includegraphics[]{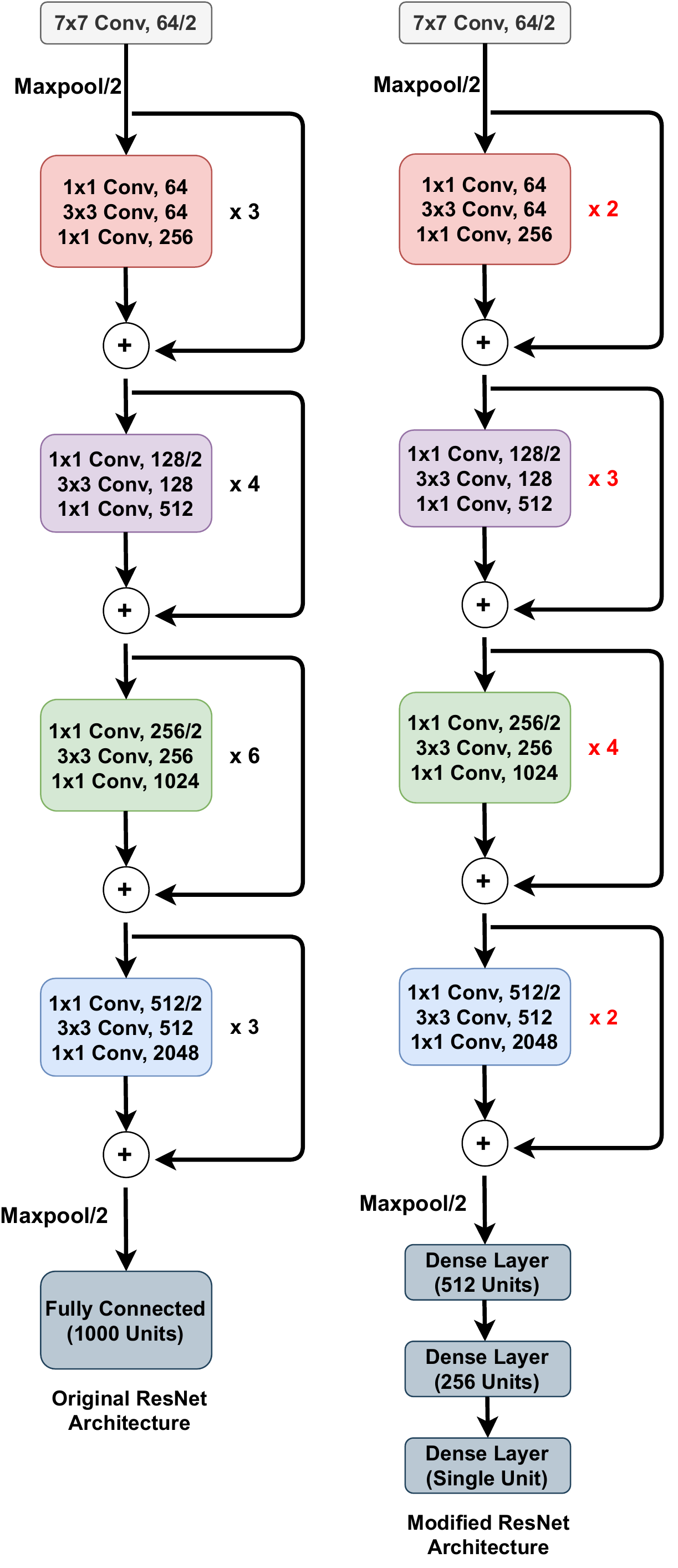}
	}}
	\caption{Block diagram of original and modified ResNet architecture}
	\label{Both_ResNets}

 \vspace{-1mm}	
 \end{figure}

The process of highlighting the driving features by introducing the encoder-decoder and channel attention impose computational overhead on the entire system while an FPAD system should take a minimum amount of time to classify the input fingerprint sample. We reduced the depth of the opted CNN architecture without tampering with its spatial properties to address the overhead imposed by the heatmap generator block. The original ResNet architecture consists of four building blocks, each having a set of three convolutional layers. In ResNet-50, the first, second, third and fourth blocks are repeated 3, 4, 6 and 3 times, respectively. In this way, the total number of convolutional layers in it is 48 ($3\times3 + 3\times4 + 3\times6 + 3\times3$). This architecture had been proposed to deal with the problem of vanishing gradient that occurs when the CNNs are trained with images that have fewer features. The skip connections between the blocks maintain the gradient and persist the parameters to learn for better classification. The depth of the Resnet can be reduced in two ways. i.e. removing the presence of the entire last block or reducing the repetitions of all the blocks.\\ 
In the first approach, the number of feature maps is reduced as we remove the last convolutional block along with its repetitions. The major disadvantage of this approach is that we get the bigger-sized feature maps at the end of the architecture resulting in decreased classification performance. We choose to use the second strategy, which minimizes the recurrence of a block resulting in less number of layers as compared with the original ResNet architecture. In this approach, there are 30 convolutional layers since the first convolution block repeats twice, the second block twice, the third block four times and the last block twice.
As we obtain feature maps at every level with the same size as the original architecture, the removal of the layers in this manner survives in the spatial attributes of feature maps. The Slim ResNet architecture's spatial dimension consists of 2048 feature maps of the size of $7 \times 7$ pixels each. Since, the input samples are resized, we get the feature maps of the size of  $16 \times 16$. The output feature maps undergo the process of pooling which results in an array of 2048 values. The downsample the output of the convolution base, and make it suitable for binary classification, three fully-connected layers with 512, 256 and 1 neurons, respectively, are added.
The original and the modified ResNet architecture are depicted in Fig. \ref{Both_ResNets}.




\section{Experimental Setup} \label{experimental results}
In this section, we discuss the different databases used in our experimental evaluation, the performance metrics for evaluation and the implementation details of our method.
\subsection{\textbf{Database}}
The performance of the proposed model is validated using LivDet 2011, 2013, 2015, 2017 and 2019 databases. Each database is prepared with multiple sensing devices. The training and testing fingerprint samples are arranged in a separate group of datasets. The details of all the utilized databases are mentioned in Table \ref{Database_details}.
\begin{table*}[!hbt]
\begin{center}
\caption{Details of the benchmark LivDet databases}
\label{Database_details}
\resizebox{0.8\textwidth}{!}{
\begin{tabular}{|l|l|c|c|l|}
\hline
\textbf{Database}                     & \textbf{Sensor}          & \textbf{Live} & \textbf{Spoof} & \textbf{Spoofing   Materials}                                                      \\ \hline
{\textbf{LivDet 2011 \cite{2011}}} & \textbf{Biometrika}      & 1000/1000     & 1000/1000      & {Ecoflex, Gelatine, Latex, Siligum, Woodglue}                       \\ \cline{2-4}
                                      & \textbf{Italdata}        & 1000/1000     & 1000/1000      &                                                                                    \\ \cline{2-5} 
                                      & \textbf{Digital Persona} & 1000/1000     & 1000/1000      & \multirow{2}{*}{Gelatine, Latex, Playdoh, Silicone, WoodGlue}                      \\ \cline{2-4}
                                      & \textbf{Sagem}           & 1000/1000     & 1000/1000      &                                                                                    \\ \hline
{\textbf{LivDet 2013 \cite{2013}}} & \textbf{Biometrika}      & 1000/1000     & 1000/1000      & {Ecoflex, Gelatine, Latex, Modsil, Woodglue}                        \\ \cline{2-4}
                                      & \textbf{Digital Persona} & 1000/1000     & 1000/1000      &                                                                                    \\ \hline
{\textbf{LivDet 2015 \cite{b34_livdet2015}}} & \textbf{Crossmatch}      & 1000/1000     & 1473/1448      & Body Double, Ecoflex, Playdoh, OOMOO, Gelatine                                     \\ \cline{2-5} 
                                      & \textbf{Digital Persona} & 1000/1000     & 1000/1500      & \multirow{3}{*}{Ecoflex, Latex, Gelatine, Woodglue, Liquid Ecoflex,   RTV}         \\ \cline{2-4}
                                      & \textbf{Greenbit}        & 1000/1000     & 1000/1500      &                                                                                    \\ \cline{2-4}
                                      & \textbf{Hi-Scan}         & 1000/1000     & 1000/1500      &                                                                                    \\ \hline
{\textbf{LivDet 2017 \cite{livdet_2017}}} & \textbf{Greenbit}        & 1000/1700     & 1200/2040      & \multirow{3}{*}{Body Double, Ecoflex, Woodglue, Gelatine, Latex,   Liquid Ecoflex} \\ \cline{2-4}
                                      & \textbf{Orcanthus}       & 1000/1700     & 1180/2018      &                                                                                    \\ \cline{2-4}
                                      & \textbf{Digital Persona} & 999/1692      & 1199/2028      &                                                                                    \\ \hline
{\textbf{LivDet 2019 \cite{b30_orru}}} & \textbf{Greenbit}        & 1000/1020     & 1200/1224      &            Body Double, Ecoflex, Woodglue, Mix1, Mix2, Liquid Ecoflex                                                                        \\ \cline{2-5} 
                                      & \textbf{Orcanthus}       & 1000/990      & 1200/1088      &   Body Double, Ecoflex, Woodglue, Mix1, Mix2, Liquid Ecoflex                                                                                 \\ \cline{2-5} 
                                      & \textbf{Digital Persona} & 1000/1099     & 1000/1224      &     Ecoflex, Gelatine, Woodglue, Latex, Mix1, Mix2, Liquid Ecoflex                                                                               \\ \hline
\end{tabular}}
\end{center}
\end{table*}
Table \ref{Database_details} describes the information of sensors, number of live and spoof samples and materials utilized for the fabrication of spoofs. The sensors including, Biometrika (hi-Scan), italdata, digital-persona, sagem, crossmatch and greenbit are optical sensors while orcanthus is a thermal sensor. The samples captured with orcanthus consist of noise and scars making them hard to classify for an FPAD model.  
\subsection{Performance Metrics}
The performance of the proposed model is measured using ISO/IEC IS 30107 criteria \cite{misc1}. The Attack Presentation Classification Error Rate (APCER) shows the percentage of misclassified spoof fingerprint images and its counterpart the Bonafide Presentation Classification Error Rate (BPCER), shows the percentage of misclassified live fingerprint images. APCER and BPCER are donated by Eq. (\ref{APCER}) and Eq. (\ref{BPCER}) respectively.
\begin{equation}
\label{APCER}
APCER = \frac{\textnormal{Number of mis-classified fake samples}}{\textnormal{Total fake samples}} \times{100}
\end{equation}

\begin{equation}
\label{BPCER}
BPCER = \frac{\textnormal{Number of mis-classified live samples}}{\textnormal{Total live samples}} \times {100}
\end{equation}

The Average classification error (ACE) is calculated by taking an average of APCER and BPCER and is used to evaluate the system's overall performance.
Equation (\ref{ACE}) represents the formulation of ACE.
\begin{equation}
\label{ACE}
ACE = \frac{APCER + BPCER}{2}
\end{equation}
The ACE is further utilized to derive the accuracy of the proposed model which is formulated as Eq. (\ref{Accuracy}).
\begin{equation}
\label{Accuracy}
Accuracy = 100 -  ACE
\end{equation}
\subsection{\textbf{Implementation Details}}
The proposed algorithm is implemented in python using the Tensorflow-Keras library. All training and testing have been done over NVIDIA TESLA P100 GPU. Each model has been trained from scratch for 250 epochs which took around 10-12 hours to converge. The learning rate and batch size are kept as 0.0001 and 8 respectively.
\section{\textbf{Experimental Results and Comparative Analysis}}
\subsection{Experimental Results}
The performance of the proposed model is validated in two different benchmark protocols, including intra-sensor and known spoof material and intra-sensor and unknown spoof material based on the arrangement of training and testing spoof samples captured with multiple devices. A description of these protocols along with the findings of the proposed method is discussed in the following subsections.
\subsubsection{\textbf{Intra-Sensor and Known Spoof Material}} In this experimental setup, the training and testing fingerprint samples are captured using the same sensing device. The spoof samples belonging to both training and testing datasets are fabricated with the same spoofing materials. LivDet 2011 and 2013 are prepared according to this setup while LivDet 2015 partially belongs to this category as two-thirds of the testing samples are captured using known spoof materials. The results on LivDet 2011 and 2013 databases are reported in Table \ref{tab: 2011_2013 intra-sensor}. Table \ref{tab: 2011_2013 intra-sensor} indicates that the proposed model attains an average BPCER of 3.50\%, APCER of 2.79\% and ACE of 3.14\% while being tested on the LivDet 2011 database. In the same protocol, the model achieves a BPCER of 0.15\%, APCER of 0.17\% and ACE of 0.16\%  while being tested on the LivDet 2013 database. The results on LivDet 2015 are reported in Table \ref{tab : 2015_Intra_Sensor} which indicates that the proposed model achieves an average BPCER of 3.23\% and APCER of 2.91\% as mentioned by the column ``APCER (Known)". 
\begin{table}[!htb]
\caption{The performance on LivDet 2011, 2013 database on intra-sensor known-material protocol}
\label{tab: 2011_2013 intra-sensor}
\begin{tabular}{|l|l|c|c|c|}
\hline
\textbf{Database} & \textbf{Sensor} & \textbf{BPCER} & \textbf{APCER} & \textbf{ACE (\%)} \\ \hline
{\textbf{LivDet 2011}} & \textbf{Biometrika} &7.1   &1.6  &4.35  \\ \cline{2-5} 
 & \textbf{Digital Persona} &1.9  &1.0  &1.45  \\ \cline{2-5} 
 & \textbf{Italdata} &3.9  &7.0  &5.45  \\ \cline{2-5} 
 & \textbf{Sagem} &1.12  &1.58  &1.35  \\ \cline{2-5} 
 & \textbf{Average} & \textbf{3.50} & \textbf{2.79} & \textbf{3.14} \\ \hline
{\textbf{LivDet 2013}} & \textbf{Biometrika} &0.15  &0.15  &0.15  \\ \cline{2-5} 
 & \textbf{Italdata} &0.15  &0.20  &0.17  \\ \cline{2-5} 
 & \textbf{Average} & \textbf{0.15} & \textbf{0.17} & \textbf{0.16} \\ \hline
\end{tabular}
\end{table}
\begin{table}[!bht]
\caption{The performance on LivDet 2015 database on intra-sensor known and unknown materials protocols}
\label{tab : 2015_Intra_Sensor}
\resizebox{0.5\textwidth}{!}{
\begin{tabular}{|l|l|c|c|c|c|}
\hline
\textbf{Database} & \textbf{Sensor} & \textbf{BPCER} & \textbf{\begin{tabular}[c]{@{}l@{}}APCER\\ (Known)\end{tabular}} & \textbf{\begin{tabular}[c]{@{}l@{}}APCER \\ (Unknown)\end{tabular}} & \textbf{ACE (\%)} \\ \hline
\multirow{5}{*}{\textbf{LivDet 2015}} & \textbf{Crossmatch} & 1.0 & 3.41  &11.06  &3.7  \\ \cline{2-6} 
 & \textbf{Digital Persona} &5.6  &3.5 &3.2  &4.49  \\ \cline{2-6} 
 & \textbf{Biometrika} & 4.0 &3.2  &5.2  &3.93  \\ \cline{2-6} 
 & \textbf{Greenbit} &2.11  &1.30  &3.6  &2.08  \\ \cline{2-6} 
 & \textbf{Average} & \textbf{3.23} & \textbf{2.91} & \textbf{5.82} & \textbf{3.55} \\ \hline
\end{tabular}}
\end{table}
\subsubsection{\textbf{Intra-Sensor and Unknown Spoof Material}}
In this experimental setup, the fingerprint samples belonging to the training and testing datasets are captured using the same sensing device however the samples belonging to the spoof category in both datasets are fabricated using different materials. Validation in this protocol, measures the robustness of the FPAD system to defend the AFRS in the real-world scenario since an intruder can present an artifact of a user's fingerprint made with newly discovered fabrication materials that are unseen to the FPAD model. LivDet 2017 and 2019 are captured in the same way as the training and testing spoof samples are fabricated from different materials. The findings of the proposed method on the aforementioned databases are reported in Table \ref{tab: intra_2017_2019}.  Table \ref{tab: intra_2017_2019} shows that the proposed model achieves an average BPCER of 4.70\%, APCER of 3.28\% and ACE of 3.92\% on the LivDet 2017 database. Similarly the proposed model classifies the live and spoof samples with an error of 4.68\% and 2.96\% respectively on LivDet 2019. The proposed method also confronts the spoof samples present in LivDet 2015 database with an average APCER of 5.82\% as mentioned by the column ``APCER (unknown)" in Table \ref{tab : 2015_Intra_Sensor}.

\begin{table}[!bht]
\caption{The performance on LivDet 2017 and 2019 databases on intra-sensor unknown materials protocol}
\label{tab: intra_2017_2019}
\begin{tabular}{|l|l|c|c|c|}
\hline
\textbf{Database} & \textbf{Sensor} & \textbf{BPCER} & \textbf{APCER} & \textbf{ACE (\%)} \\ \hline
\multirow{4}{*}{\textbf{LivDet 2017}} & \textbf{Digital Persona} &5.14  &3.4  &4.2  \\ \cline{2-5} 
 & \textbf{Orcanthus} &3.36  &2.86  &3.09  \\ \cline{2-5} 
 & \textbf{Greenbit} &5.59  &3.58  &4.49  \\ \cline{2-5} 
 & \textbf{Average} & \textbf{4.70} & \textbf{3.28} & \textbf{3.93} \\ \hline
\multirow{4}{*}{\textbf{LivDet 2019}} & \textbf{Digital Persona} &7.6  &7  &7.3  \\ \cline{2-5} 
 & \textbf{Greenbit} &5.3  &1.23  &3.08  \\ \cline{2-5} 
 & \textbf{Orcanthus} &1.12  &0.65  &0.87  \\ \cline{2-5} 
 & \textbf{Average} & \textbf{4.67} & \textbf{2.95} & \textbf{3.75} \\ \hline
\end{tabular}
\end{table}
\subsubsection{\textbf{Discussion}}
The properties of the live and spoof fingerprint samples differ due to the lack of moisture in the spoof. Apart from that, the spoof samples include noise, scars and uneven width of ridges and valleys that are introduced during the fabrication process. These abnormalities are emphasized by the proposed heatmap generator which plays an important role in the detection of PAs. The findings of the proposed method are compared with existing methods tested on benchmark databases which are mentioned in the following subsection.
\subsection{\textbf{Comparative Analysis \label{comparative analysis}}} The findings of the proposed method, are compared with state-of-the-art approaches in several benchmark settings. A detailed comparative analysis is given in the following subsections.
\subsubsection{\textbf{Comparison with existing methods on LivDet 2011 database}}
The performance of the proposed model is compared with state-of-the-art methods tested on LivDet 2011 database which is mentioned in Table \ref{comp_2011}. As per Table \ref{comp_2011}, the proposed method outperforms the methods discussed in \cite{xia_2}, \cite{b1_dubey}, \cite{b24_yuan1}, \cite{b40_jian}, \cite{b15_chugh2}, \cite{b5_sharma1}, \cite{b_Gragnaniello}, \cite{Nogueira} over the fingerprint samples collected with biometrika, digital-persona and sagem sensors. The spoof fingerprint samples in this database were obtained using the cooperative spoofing approach, resulting in the development of efficient spoof samples that can readily deceive a CNN-based FPAD model. The suggested heatmap generator emphasizes the presence of moisture in the input fingerprint data. As a result, spoof samples lack this feature and are easily spotted by the classifier. This advantage elevates the suggested technique over handcrafted features-based and deep CNN-based FPAD approaches. The proposed method attains overall classification accuracy of 96.86\%.
\begin{table}[!bht]
\caption{Comparison with state-of-the-art methods on LivDet 2011 in intra-sensor protocol } 
\label{comp_2011}
\resizebox{0.50\textwidth}{!}{
\begin{tabular}{|p{2.5 cm}|c|c|c|c|c|}
\hline
\textbf{Method} & \textbf{\begin{tabular}[c]{@{}c@{}}Accuracy\\(Biometrika)\end{tabular}}& \textbf{\begin{tabular}[c]{@{}c@{}}Accuracy\\(Digital Persona)\end{tabular}}& \textbf{\begin{tabular}[c]{@{}c@{}}Accuracy\\(Italdata)\end{tabular}}& \textbf{\begin{tabular}[c]{@{}c@{}}Accuracy\\(Sagem)\end{tabular}}& \multicolumn{1}{l|}{\textbf{Avg.}} \\ \hline
Xia et al. \cite{xia_2}           & 93.55                                   & 96.2                                          & 88.25                                  & 96.66                               & 93.37                                  \\ \hline
 Dubey et al. \cite{b1_dubey}         & 92.11                                   & 93.75                                         & 91.9                                   & 94.64                               & 93.1                                   \\ \hline
Yuan et al. \cite{b24_yuan1}          & 97.05                                   & 88.94                                         & 97.8                                   & 92.01                               & 93.82                                  \\ \hline
Gragnaniello et al. \cite{b_Gragnaniello}  & 93.1                                    & 92.00                                         & 87.35                                  & 96.35                               & 92.2                                   \\ \hline
Nogueira et al. \cite{Nogueira}      & 91.8                                    & 98.1                                          & 94.91                                  & 95.36                               & 95.04                                  \\ \hline
Yuan et al. \cite{b24_yuan1}          & 90.08                                   & 98.65                                         & 87.65                                  & 97.1                                & 93.55                                  \\ \hline
 Jian et al. \cite{b40_jian}          & 95.75                                   & 98.4                                          & 94.1                                   & 96.83                               & 96.27                                     \\ \hline Sharma et al. \cite{b5_sharma1}                     &  92.7                                  &     94.4                                     &       88.6                           &    93.3                            &             92.25                      \\ \hline
Chugh et al. \cite{b15_chugh2}                     & 98.76                                   & 98.39                                         & 97.55                                  & 98.61                               & 98.33                                  \\ 
\hline

                                       \textbf{EXPRESSNET}          & \textbf{95.65}                          & \textbf{98.55}                                & \textbf{94.55}                         & \textbf{98.65}                      & \textbf{96.86}                         \\\hline
\end{tabular}}
\end{table}
\subsubsection{\textbf{Comparison with existing methods on LivDet 2013 database}}
The findings of the proposed method are compared with the method tested on the LivDet 2013 database. This database is captured using the non-cooperative method of spoofing in which the latent fingerprints left on the glass, wood, or other smooth surface are used to fabricate the spoofs. This process adds a significant amount of noise, scars and other irregularities to the spoofs which are highlighted by the heatmap generator. Table \ref{comp_2013} shows a detailed comparison of the proposed method's performance with state-of-the-art methods validated on the LivDet 2013 database. It is evident that the proposed methods perform better as compared with the method discussed in \cite{b24_yuan1}, \cite{b40_jian}, \cite{b31_zhang}, \cite{b41_park2}, \cite{b42_Gottschlich}, \cite{b43_johnson}, \cite{b44_yuan3}, \cite{b45_jung}, \cite{b4_uliyan}, \cite{Nogueira}, \cite{b10_anusha}, and \cite{b15_chugh2}, while being tested on dataset captured with biometrika and italdata sensors. 

\begin{table}[!bht]
\caption{Comparison with state-of-the-art methods on LivDet 2013 in intra-sensor protocol }    
\label{comp_2013}
\resizebox{0.50\textwidth}{!}{
\begin{tabular}{ |p{2.5 cm}|c|c|c|}
\hline
\textbf{Method}                                                                                                    & \textbf{\begin{tabular}[c]{@{}c@{}}Accuracy\\(Biometrika)\end{tabular}} & \textbf{\begin{tabular}[c]{@{}c@{}}Accuracy\\(Italdata)\end{tabular}} & \multicolumn{1}{l|}{\textbf{Avg.}} \\ \hline
Yuan et al. \cite{b24_yuan1}                                                                                                       & 96.45                                             & 97.65                                           & 97.05                                 \\ \hline
 Jian et al. \cite{b40_jian}                                                                                                           & 99.25                                             & 99.40                                           & 99.32                                 \\ \hline
Zhang et al. \cite{b31_zhang}                                                                                                   & 99.53                                             & 96.99                                           & 98.26                                 \\ \hline
Park et al. \cite{b41_park2}                                                                                          & 99.15                                             & 98.75                                           & 98.95                                 \\ \hline
Gottschlich et al. \cite{b42_Gottschlich}                      & 96.10                                             & 98.30                                           & 97.0                                  \\ \hline
Johnson et al. \cite{b43_johnson}                                                                  & 98.0                                              & 98.4                                            & 98.20                                  \\ \hline
Yuan et al. \cite{b44_yuan3}                                                                            & 95.65                                             & 98.6                                            & 97.12                                 \\ \hline
Jung et al. \cite{b45_jung}                                                                                  & 94.12                                             & 97.92                                           & 96.02                                 \\ \hline
Uliyan et al. \cite{b4_uliyan}                                                                         & 96.0                                              & 94.50                                           & 95.25                                 \\ \hline
Nogueira et al. \cite{Nogueira}                                                                                  & 99.20                                             & 97.7                                            & 98.45                                 \\ \hline
Chugh et al. \cite{b15_chugh2}                                                                                  & 99.80                                             & 99.70                                            & 99.75                                 \\ \hline
Anusha et al. \cite{b10_anusha}                                                                                  & 99.76                                             & 99.68                                            & 99.72                                 \\ \hline
\textbf{EPRESSNET}                                                                                           & \textbf{99.85}                                    & \textbf{99.83}                                  & \textbf{99.84}                       \\ \hline
\end{tabular}
}
\end{table}

\subsubsection{\textbf{Comparison with existing methods on LivDet 2015 database}}
The LivDet 2015 database is composed of the spoof samples captured with known and unknown spoofing materials. A detailed comparison mentioned in Table \ref{tab: Comparison_Intra_2015} clearly indicates that the classification performance of the proposed method is better than the method discussed in \cite{b41_park2}, \cite{b5_sharma1}, \cite{b34_livdet2015}, \cite{b4_uliyan} and \cite{b23_kim}. The heatmap generator finds discriminating features that result in better classification accuracy of the classifier than state-of-the-art deep CNN-based approaches.

\begin{table}[!hbt]
 \caption{Comparison with state-of-the-art methods on LivDet 2015 in intra-sensor protocol }
\label{tab: Comparison_Intra_2015} 
\resizebox{0.50\textwidth}{!}{
\begin{tabular}{|p{2.5cm}|c|c|c|c|c|}
\hline
 \textbf{Method}                    & \textbf{\begin{tabular}[c]{@{}c@{}}Accuracy\\(Crossmatch)\end{tabular}} & \textbf{\begin{tabular}[c]{@{}c@{}}Accuracy \\(Greenbit)\end{tabular}} & \textbf{\begin{tabular}[c]{@{}c@{}}Accuracy\\(Digital Persona)\end{tabular}} & \textbf{\begin{tabular}[c]{@{}c@{}}Accuracy \\ (Biometrika)\end{tabular}} & \textbf{Avg.}                      \\ \hline
                                     Park et al.\cite{b41_park2}                                                                                     & 99.63                                                                    & 97.30                                                                   & 91.5                                                                         & 95.9                                                                       & 96.08                        \\ \hline

Sharma et al. \cite{b5_sharma1}                                                                                    & 98.07                                                                    & 95.7                                                                    & 94.16                                                                        & 95.22                                                                      & 95.78                        \\ \hline

Zhang et al. \cite{b31_zhang}                                                                                      & 97.01                                                                    & 97.81                                                                   & 95.42                                                                        & 97.02                                                                      & 96.82                        \\ \hline
Jung et al. \cite{b47_jung2}                                                                                       & 98.60                                                                    & 96.20                                                                   & 90.50                                                                        & 95.80                                                                      & 95.27                        \\ \hline
LivDet 2015 Winner \cite{b34_livdet2015}                                                                          & 98.10                                                                    & 95.40                                                                   & 93.72                                                                        & 94.36                                                                      & 95.39                        \\ \hline
Uliyan et al. \cite{b4_uliyan}                                                                                     & 95.00                                                                    & -                                                                       & -                                                                            & -                                                                          & 95.00                        \\ \hline

Kim et al. \cite{b23_kim}                                                                                         & -                                                                        & -                                                                       & -                                                                            & -                                                                          & 86.39                        \\ \hline
             {\textbf{EXPRESSNET}}                                                              & {\textbf{96.30}}                                    & { \textbf{97.92}}                                   & { \textbf{95.51}}                                        & { \textbf{96.12}}                                      & {\textbf{96.45}} \\ \hline
\end{tabular}}

\end{table}

\subsubsection{\textbf{Comparison with existing methods on LivDet 2017 database}}
The performance of the proposed method is also compared with state-of-the-art methods tested on LivDet 2017 database. The training and testing spoof samples captured in this database are fabricated using different spoofing materials which makes it more challenging for an FPAD model to classify. However, the fabrication materials available for the spoofing, do not resemble the moisture present in the live fingerprint samples. The proposed method is able to find the discriminating features with the help of the heatmap generator. Table \ref{tab: Comparison_Intra_2017} shows that the proposed method performs better than the method discussed in \cite{b14_chugh1}, \cite{b15_chugh2}, \cite{b31_zhang} and \cite{GONZALEZ-SOLER} while being tested on the fingerprint samples captured with orcanthus and digital persona. The proposed method also outperforms the aforementioned methods with an average classification accuracy of 96.07\%. This comparison reveals that the heatmap generator can produce a heatmap with discriminating information regardless of the material used for fabrication.
\begin{table}[h]
\begin{center}
\caption{Comparison with state-of-the-art methods on LivDet 2017 database in intra-sensor protocol}
\label{tab: Comparison_Intra_2017}
\resizebox{0.50\textwidth}{!}{
\begin{tabular}{|p{2.1 cm} |c| c| c| c|}
\hline
\textbf{Method} & \textbf{\begin{tabular}[c]{@{}c@{}}Accuracy \\(Orcanthus)\end{tabular}}& \textbf{\begin{tabular}[c]{@{}c@{}}Accuracy \\(Digital Persona)\end{tabular}} & \textbf{\begin{tabular}[c]{@{}c@{}}Accuracy \\(Greenbit)\end{tabular}} & \multicolumn{1}{l|}{\textbf{Avg.}} \\ \hline
Chugh et al. \cite{b14_chugh1}              & 95.01                                           &   95.20                                                 & 97.42                                            & 95.88                                 \\ \hline
Chugh et al. \cite{b15_chugh2}              & 94.51                                            & 95.12                                                  & 96.68                                            & 95.43                                 \\ \hline
Zhang et al. \cite{b31_zhang}              & 93.93                                            & 92.89                                                  & 95.20                                            & 94.00                                 \\ \hline
Gonzalez et al. \cite{GONZALEZ-SOLER}              & 94.38                                            & 95.08                                                  & 94.54                                            & 94.66                                 \\ \hline
                 \textbf{EXPRESSNET}           & \textbf{96.95}                                   & \textbf{95.80}                                         & \textbf{95.51}                                   & \textbf{96.07}                        \\ \hline
\end{tabular}
}
\end{center}
\end{table}
\subsubsection{\textbf{Comparison with existing methods on LivDet 2019 database}}
Table \ref{tab: Comparison_Intra_2019} reports a comparison of the proposed model's findings with state-of-the-art methods tested on the LivDet 2019 database. It shows that the proposed method outperforms the method discussed in \cite{b15_chugh2} as well as the participating FPAD algorithms i.e., JungCNN, JWL LivDet, ZJUT DET while being tested on the samples collected with orcanthus and digital persona sensors. The proposed method also outperforms the aforementioned methods in terms of average classification accuracy. \\
The comparative analysis of the performance of the proposed method on various LivDet databases indicates that it consistently performs better regardless of the sensors in the intra-sensor paradigm of FPAD whether the spoof samples are fabricated using known or unknown materials. The possession of the heatmap generator enables the classifier to learn better as compared with traditional CNN-based approaches.

\begin{table}[!bht]
\begin{center}
\caption{Comparison with state-of-the-art methods on LivDet 2019 database in intra-sensor protocol}
\label{tab: Comparison_Intra_2019}
\resizebox{0.50\textwidth}{!}{
\begin{tabular}{|p{2.1 cm} |c| c| c| c|}
\hline
\textbf{Method} & \textbf{\begin{tabular}[c]{@{}c@{}}Accuracy \\(Orcanthus)\end{tabular}}& \textbf{\begin{tabular}[c]{@{}c@{}}Accuracy \\(Digital Persona)\end{tabular}} & \textbf{\begin{tabular}[c]{@{}c@{}}Accuracy \\(Greenbit)\end{tabular}} & \multicolumn{1}{l|}{\textbf{Avg.}} \\ \hline

Jung CNN \cite{b30_orru}            & 99.13                                           &81.23                                                   &  99.06                                           & 93.14                                \\ \hline
Chugh et al. \cite{b15_chugh2}              &    97.50                                        &     83.64                                              &99.73                                             & 93.62                                \\ \hline
JWL LivDet \cite{b30_orru}              &      97.45                                      &       88.86                                            & 99.20                                            &   95.17                              \\ \hline
ZJUT Det A \cite{b30_orru}              &       97.50                                     &      88.77                                             &    99.20                                         &              95.16                   \\ \hline

                 \textbf{EXPRESSNET}           & \textbf{99.16}                                   & \textbf{92.70}                                         & \textbf{96.92}                                   & \textbf{96.27}                        \\ \hline
\end{tabular}
}
\end{center}
\end{table}
\subsection{Evaluation of EXPRESSNET in High-Security Systems}
\begin{figure*}[h!]
	\centering
	\resizebox{0.70\textwidth}{!}{	
	\subfigure
	{	
		\includegraphics[width=0.30\textwidth]{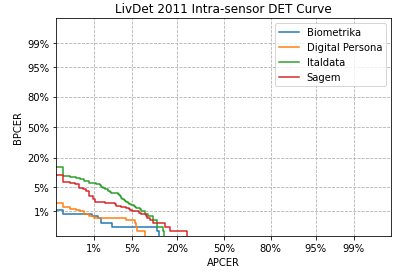}
		\hspace{4 mm}
		\includegraphics[width=0.30\textwidth]{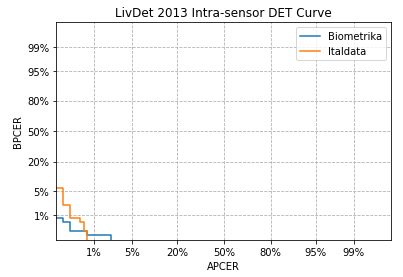}
	}}
	\resizebox{0.70\textwidth}{!}{
	\subfigure
	{
		\includegraphics[width=0.30\textwidth]{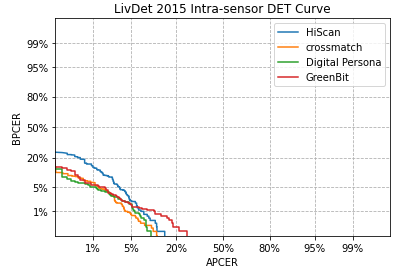}
		\hspace{4 mm}
		\includegraphics[width=0.30\textwidth]{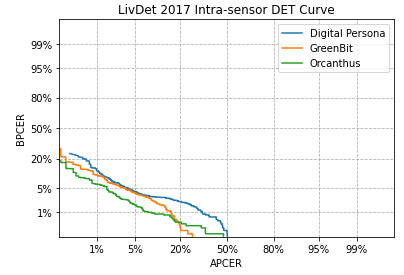}
	}}
	\resizebox{0.40\textwidth}{!}{
	\subfigure
	{	
		\includegraphics[width=0.30\textwidth]{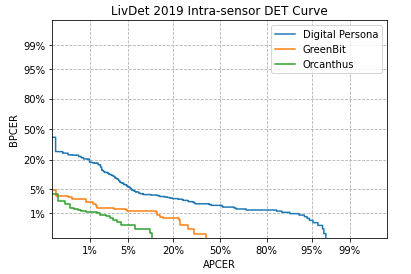}
}}

	\caption{Detection Error Trade-off (DET) curves for LivDet 2011, 2013, 2015, 2017 and 2019 databases}
	\label{DET_Cur}
	\vspace{-2mm}
\end{figure*}

\vspace{-0.5 mm}
An FPAD model is to be tested for its performance in high-security systems too as its main objective is not only to achieve the minimum APCER, BPCER and ACE. In this paper, we have reported the findings of the proposed model using the Detection Error Trade-off (DET) curve. A DET curve is a graphical representation of error rates achieved by a binary classification system by adjusting the value of the classification threshold. We have reported the DET curves for all the datasets of LivDet 2011, 2013, 2015, 2017 and 2019 databases which are depicted in Fig. \ref{DET_Cur}. In the Fig. \ref{DET_Cur}, it can be observed that the proposed model attains the BPCER of less than 1\% to retain the APCER of 1\% on biometrika and digital persona sensors of the LivDet 2011 database while it is less than 5\% and 22\% on sagem and italdata sensors of the same database. On LivDet 2013, the proposed model achieves a BPCER of less than 1\% to maintain the APCER of 1\% on biometrika and italdata sensors. Similarly, the proposed model is able to achieve a BPCER of less than 5\% to gain the APCER of 1\% while being tested on crossmatch, digital-persona and greenbit sensors of the LivDet 2015 database. LivDet 2017 and 2019 in which the testing spoof samples are captured using unknown spoof materials, the model is able to retain the BPCER in the range of 5\% - 17\% on Livdet 2017 database. In the same way, the model retains the BPCER of less than 5\% on orcanthus and greenbit sensors of the LivDet 2019 database.

\subsection{\textbf{Processing Time}}
The processing time of an FPAD model is considered the amount of time it takes to find whether the input fingerprint sample is live or spoof. This time is supposed to be minimum as the sample has to undergo the process of verification after the detection of its liveness. The proposed model, $EXPRESSNET$, takes the classification time of 300 milliseconds and 20 milliseconds on Intel(R) Core(TM) i5-6500 CPU @ 3.20GHz $6^{th}$ generation processor, and Nvidia TESLA P00 respectively, to classify a single fingerprint image. The less amount of classification time makes it suitable for the AFRS in real-time applications.
\section{Conclusion \label{conclusion}}
AFRS deployed in various security and commercial applications can be deceived by PAs. This paper presents an FPAD mechanism that has shown the capability of detecting spoofs when they are created using cooperative, or non-cooperative methods of spoofing as well as using known and unknown fabrication materials. Existing handcrafted and deep learning-based methods are insufficient in detecting PAs while being tested in the aforementioned scenarios. One of the possible reasons behind this is the lack of feature extraction capability of CNN-based methods due to the limited amount of discriminating information present in the input fingerprint samples. In this paper, a novel end-to-end model is presented which first converts the input fingerprint sample into a heatmap that represents the most informative part of it. It also finds what is important in the input sample. The generated heatmap is then fed to the standard modified CNN classifier for classification. We have tested the method on benchmark databases in different experimental settings. The efficacy of the proposed models is compared with state-of-the-art methods including statistical and handcrafted feature-based methods, perspiration and pore-based methods and deep learning-based methods. In future, we will explore the capability of the proposed model for cross-sensor and cross-database validation on benchmark fingerprint databases.

\bibliographystyle{IEEE}
\bibliography{references}

\begin{thebibliography}{10}
\providecommand{\url}[1]{\texttt{#1}}
\providecommand{\urlprefix}{URL }
\providecommand{\doi}[1]{https://doi.org/#1}

\bibitem{misc1}
30107-3:2017(en), I.: Information technology — biometric presentation attack
  detection — part 3: Testing and reporting (2017)

\bibitem{b19_abhyankar}
Abhyankar, A., Schuckers, S.: Integrating a wavelet based perspiration liveness
  check with fingerprint recognition. Pattern Recognition  \textbf{42},
  452--464 (03 2009)

\bibitem{b10_anusha}
Anusha, B., Banerjee, S., Chaudhuri, S.: Defraudnet:end2end fingerprint spoof
  detection using patch level attention. In: 2020 IEEE Winter Conference on
  Applications of Computer Vision (WACV). pp. 2684--2693. IEEE Computer
  Society, Los Alamitos, CA, USA (mar 2020)

\bibitem{b26_arora}
Arora, S.: Fingerprint spoofing detection to improve customer security in
  mobile financial applications using deep learning. Arabian Journal for
  Science and Engineering  \textbf{45} (10 2019)

\bibitem{b7_choi1}
Choi, H., Kang, R., Choi, K., Teoh, A., Kim, J.: Fake-fingerprint detection
  using multiple static features. Optical Engineering - OPT ENG  \textbf{48}
  (04 2009)

\bibitem{b15_chugh2}
Chugh, T., Cao, K., Jain, A.K.: Fingerprint spoof buster: Use of
  minutiae-centered patches. IEEE Transactions on Information Forensics and
  Security  \textbf{13}(9),  2190--2202 (2018)

\bibitem{b14_chugh1}
Chugh, T., Jain, A.K.: Fingerprint spoof detector generalization. IEEE
  Transactions on Information Forensics and Security  \textbf{16},  42--55
  (2021)

\bibitem{imagenet}
Deng, J., Dong, W., Socher, R., Li, L.J., Li, K., Fei-Fei, L.: Imagenet: A
  large-scale hierarchical image database. In: 2009 IEEE Conference on Computer
  Vision and Pattern Recognition. pp. 248--255 (2009)

\bibitem{mnist}
Deng, L.: The mnist database of handwritten digit images for machine learning
  research [best of the web]. IEEE Signal Processing Magazine  \textbf{29}(6),
  141--142 (2012)

\bibitem{b27_derakshini}
Derakhshani, R., Schuckers, S.A., Hornak, L.A., O'Gorman, L.: Determination of
  vitality from a non-invasive biomedical measurement for use in fingerprint
  scanners. Pattern Recognition  \textbf{36}(2),  383--396 (2003)

\bibitem{b1_dubey}
Dubey, R.K., Goh, J., Thing, V.L.L.: Fingerprint liveness detection from single
  image using low-level features and shape analysis. IEEE Transactions on
  Information Forensics and Security  \textbf{11}(7),  1461--1475 (2016)

\bibitem{EXP6}
Erhan, D., Bengio, Y., Courville, A., Vincent, P.: Visualizing higher-layer
  features of a deep network. Technical Report, Univeristé de Montréal  (01
  2009)

\bibitem{b16_espinoza}
Espinoza, M., Champod, C.: Using the number of pores on fingerprint images to
  detect spoofing attacks. In: 2011 International Conference on Hand-Based
  Biometrics. pp.~1--5 (2011)

\bibitem{b12_ghiani}
Ghiani, L., Hadid, A., Marcialis, G.L., Roli, F.: Fingerprint liveness
  detection using binarized statistical image features. In: 2013 IEEE Sixth
  International Conference on Biometrics: Theory, Applications and Systems
  (BTAS). pp.~1--6 (2013)

\bibitem{2013}
Ghiani, L., Yambay, D., Mura, V., Tocco, S., Marcialis, G.L., Roli, F.,
  Schuckcrs, S.: Livdet 2013 fingerprint liveness detection competition 2013.
  In: 2013 International Conference on Biometrics (ICB). pp.~1--6 (2013)

\bibitem{GONZALEZ-SOLER}
González-Soler, L.J., Gomez-Barrero, M., Chang, L., Pérez-Suárez, A., Busch,
  C.: Fingerprint presentation attack detection based on local features
  encoding for unknown attacks. IEEE Access  \textbf{9},  5806--5820 (2021)

\bibitem{b42_Gottschlich}
Gottschlich, C., Marasco, E., Yang, A.Y., Cukic, B.: Fingerprint liveness
  detection based on histograms of invariant gradients. In: IEEE International
  Joint Conference on Biometrics. pp.~1--7 (2014)

\bibitem{b_Gragnaniello}
Gragnaniello, D., Poggi, G., Sansone, C., Verdoliva, L.: Fingerprint liveness
  detection based on weber local image descriptor. In: 2013 IEEE Workshop on
  Biometric Measurements and Systems for Security and Medical Applications. pp.
  46--50 (2013)

\bibitem{resnet}
He, K., Zhang, X., Ren, S., Sun, J.: Deep residual learning for image
  recognition. In: 2016 IEEE Conference on Computer Vision and Pattern
  Recognition (CVPR). pp. 770--778 (2016)

\bibitem{b40_jian}
Jian, W., Zhou, Y., Liu, H.: Densely connected convolutional network optimized
  by genetic algorithm for fingerprint liveness detection. IEEE Access
  \textbf{9},  2229--2243 (2021)

\bibitem{b43_johnson}
Johnson, P., Schuckers, S.: Fingerprint pore characteristics for liveness
  detection. In: 2014 International Conference of the Biometrics Special
  Interest Group (BIOSIG). pp.~1--8 (2014)

\bibitem{b47_jung2}
Jung, H.Y., Heo, Y.: Fingerprint liveness map construction using convolutional
  neural network. Electronics Letters  \textbf{54} (03 2018)

\bibitem{b45_jung}
Jung, H.Y., Heo, Y.S., Lee, S.: Fingerprint liveness detection by a
  template-probe convolutional neural network. IEEE Access  \textbf{7},
  118986--118993 (2019)

\bibitem{reference1}
Khan, N., Efthymiou, M.: The use of biometric technology at airports: The case
  of customs and border protection (cbp). International Journal of Information
  Management Data Insights  \textbf{1}(2),  100049 (2021)

\bibitem{b23_kim}
Kim, W.: Fingerprint liveness detection using local coherence patterns. IEEE
  Signal Processing Letters  \textbf{24}(1),  51--55 (2017)

\bibitem{EXP7}
Lapuschkin, S., Binder, A., Montavon, G., Müller, K.R., Samek, W.: Analyzing
  classifiers: Fisher vectors and deep neural networks. In: 2016 IEEE
  Conference on Computer Vision and Pattern Recognition (CVPR). pp. 2912--2920
  (2016)

\bibitem{b25_marasco}
Marasco, E., Sansone, C.: Combining perspiration- and morphology-based static
  features for fingerprint liveness detection. Pattern Recognition Letters
  \textbf{33}(9),  1148--1156 (2012)

\bibitem{b17_marcialis}
Marcialis, G.L., Roli, F., Tidu, A.: Analysis of fingerprint pores for vitality
  detection. In: 2010 20th International Conference on Pattern Recognition. pp.
  1289--1292 (2010)

\bibitem{b34_livdet2015}
Mura, V., Ghiani, L., Marcialis, G.L., Roli, F., Yambay, D.A., Schuckers, S.A.:
  Livdet 2015 fingerprint liveness detection competition 2015. In: 2015 IEEE
  7th International Conference on Biometrics Theory, Applications and Systems
  (BTAS). pp.~1--6 (2015)

\bibitem{Nogueira}
Nogueira, R.F., de~Alencar~Lotufo, R., Campos~Machado, R.: Fingerprint liveness
  detection using convolutional neural networks. IEEE Transactions on
  Information Forensics and Security  \textbf{11}(6),  1206--1213 (2016)

\bibitem{b30_orru}
Orrù, G., Casula, R., Tuveri, P., Bazzoni, C., Dessalvi, G., Micheletto, M.,
  Ghiani, L., Marcialis, G.: Livdet in action - fingerprint liveness detection
  competition 2019. pp.~1--6 (06 2019)

\bibitem{b41_park2}
Park, E., Cui, X., Kim, W., Kim, H.: End-to-end fingerprints liveness detection
  using convolutional networks with gram module  (03 2018)

\bibitem{b11}
park, Y., Jang, U., Lee, E.C.: Statistical anti-spoofing method for fingerprint
  recognition. Soft Computing  \textbf{22}, ~1--7 (07 2018)

\bibitem{EXP2}
Ras, G., van Gerven, M., Haselager, P.: Explanation Methods in Deep Learning:
  Users, Values, Concerns and Challenges, pp. 19--36. Springer International
  Publishing, Cham (2018)

\bibitem{EXP5}
Ras, G., Xie, N., van Gerven, M., Doran, D.: Explainable deep learning: A field
  guide for the uninitiated. J. Artif. Int. Res.  \textbf{73} (may 2022)

\bibitem{b2_rattani1}
Rattani, A., Ross, A.: Automatic adaptation of fingerprint liveness detector to
  new spoof materials. In: IEEE International Joint Conference on Biometrics.
  pp.~1--8 (2014)

\bibitem{b3_rattani2}
Rattani, A., Scheirer, W.J., Ross, A.: Open set fingerprint spoof detection
  across novel fabrication materials. IEEE Transactions on Information
  Forensics and Security  \textbf{10}(11),  2447--2460 (2015)

\bibitem{EXP3}
Samek, W., Wiegand, T., Müller, K.R.: Explainable artificial intelligence:
  Understanding, visualizing and interpreting deep learning models. ITU
  Journal: ICT Discoveries - Special Issue 1 - The Impact of Artificial
  Intelligence (AI) on Communication Networks and Services  \textbf{1},  1--10
  (10 2017)

\bibitem{b37_deepika}
Sharma, D., Selwal, A.: Hyfipad: a hybrid approach for fingerprint presentation
  attack detection using local and adaptive image features. The Visual Computer
   (06 2021)

\bibitem{b5_sharma1}
Sharma, R., Dey, S.: Fingerprint liveness detection using local quality
  features. The Visual Computer  \textbf{35} (10 2019)

\bibitem{Spinoulas}
Spinoulas, L., Mirzaalian, H., Hussein, M.E., AbdAlmageed, W.: Multi-modal
  fingerprint presentation attack detection: Evaluation on a new dataset. IEEE
  Transactions on Biometrics, Behavior, and Identity Science  \textbf{3}(3),
  347--364 (2021)

\bibitem{EXP1}
Tavanaei, A.: Embedded encoder-decoder in convolutional networks towards
  explainable {AI}. CoRR  \textbf{abs/2007.06712} (2020)

\bibitem{b4_uliyan}
Uliyan, D.M., Sadeghi, S., Jalab, H.A.: Anti-spoofing method for fingerprint
  recognition using patch based deep learning machine. Engineering Science and
  Technology, an International Journal  \textbf{23}(2),  264--273 (2020)

\bibitem{cifar}
Wu, C., Li, Y., Zhao, Z., Liu, B.: Research on image classification method of
  features of combinatorial convolution. Journal of Ambient Intelligence and
  Humanized Computing  \textbf{11} (07 2020)

\bibitem{b13_xia1}
Xia, Z., Lv, R., Zhu, Y., Ji, P., Sun, H., Shi, Y.Q.: Fingerprint liveness
  detection using gradient-based texture features. Signal, Image and Video
  Processing  \textbf{11} (02 2017)

\bibitem{xia_2}
Xia, Z., Yuan, C., Lv, R., Sun, X., Xiong, N.N., Shi, Y.Q.: A novel weber local
  binary descriptor for fingerprint liveness detection. IEEE Transactions on
  Systems, Man, and Cybernetics: Systems  \textbf{50}(4),  1526--1536 (2020)

\bibitem{2011}
Yambay, D., Ghiani, L., Denti, P., Marcialis, G.L., Roli, F., Schuckers, S.:
  Livdet 2011 — fingerprint liveness detection competition 2011. In: 2012 5th
  IAPR International Conference on Biometrics (ICB). pp. 208--215 (2012)

\bibitem{livdet_2017}
Yambay, D., Schuckers, S., Denning, S., Sandmann, C., Bachurinski, A., Hogan,
  J.: Livdet 2017 - fingerprint systems liveness detection competition. In:
  2018 IEEE 9th International Conference on Biometrics Theory, Applications and
  Systems (BTAS). pp.~1--9 (2018)

\bibitem{b24_yuan1}
Yuan, C., Sun, X., Wu, Q.M.: Difference co-occurrence matrix using bp neural
  network for fingerprint liveness detection. Soft Comput.  \textbf{23}(13),
  5157–5169 (jul 2019)

\bibitem{b44_yuan3}
Yuan, C., Xia, Z., Jiang, L., Cao, Y., Jonathan~Wu, Q.M., Sun, X.: Fingerprint
  liveness detection using an improved cnn with image scale equalization. IEEE
  Access  \textbf{7},  26953--26966 (2019)

\bibitem{EXP4}
Zeiler, M.D., Fergus, R.: Visualizing and understanding convolutional networks.
  In: Fleet, D., Pajdla, T., Schiele, B., Tuytelaars, T. (eds.) Computer Vision
  -- ECCV 2014. pp. 818--833. Springer International Publishing, Cham (2014)

\bibitem{b46_zhang2}
Zhang, Y., Pan, S., Zhan, X., Li, Z., Gao, M., Gao, C.: Fldnet: Light dense cnn
  for fingerprint liveness detection. IEEE Access  \textbf{8},  84141--84152
  (2020)

\bibitem{b31_zhang}
Zhang, Y., Shi, D., Zhan, X., Cao, D., Zhu, K., Li, Z.: Slim-rescnn: A deep
  residual convolutional neural network for fingerprint liveness detection.
  IEEE Access  \textbf{7},  91476--91487 (2019)

\end{thebibliography}
\vspace{-1.5 cm}
\begin{IEEEbiography}[{\includegraphics[width=1in,height=1.25in,clip,keepaspectratio]{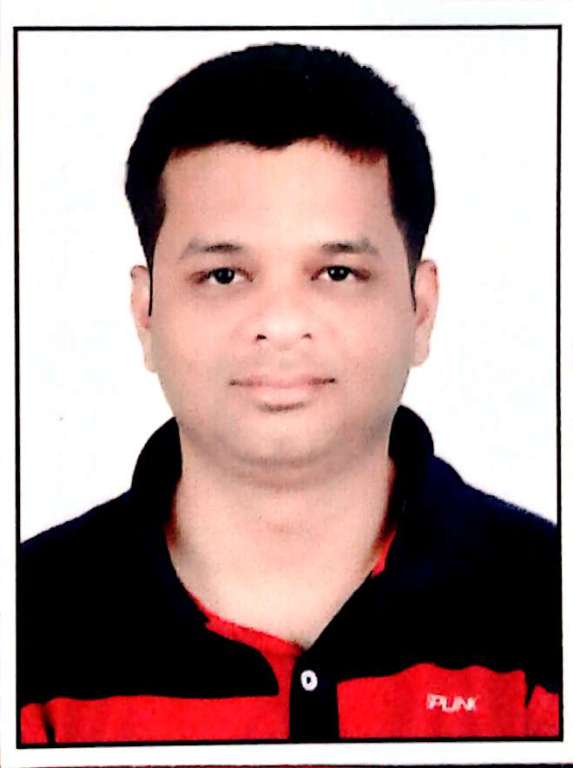}}]{Anuj Rai} received his M.Tech. degree in Computer Technology and Applications from National Institute of Technical Teachers Training and Research Bhopal, India, in 2015. He is currently pursuing his Ph.D. in Computer Science and Engineering department at Indian Institute of Technology Indore.
\end{IEEEbiography}
\vspace{-1 cm}
\begin{IEEEbiography}[{\includegraphics[width=1in,height=1.25in, clip,keepaspectratio]{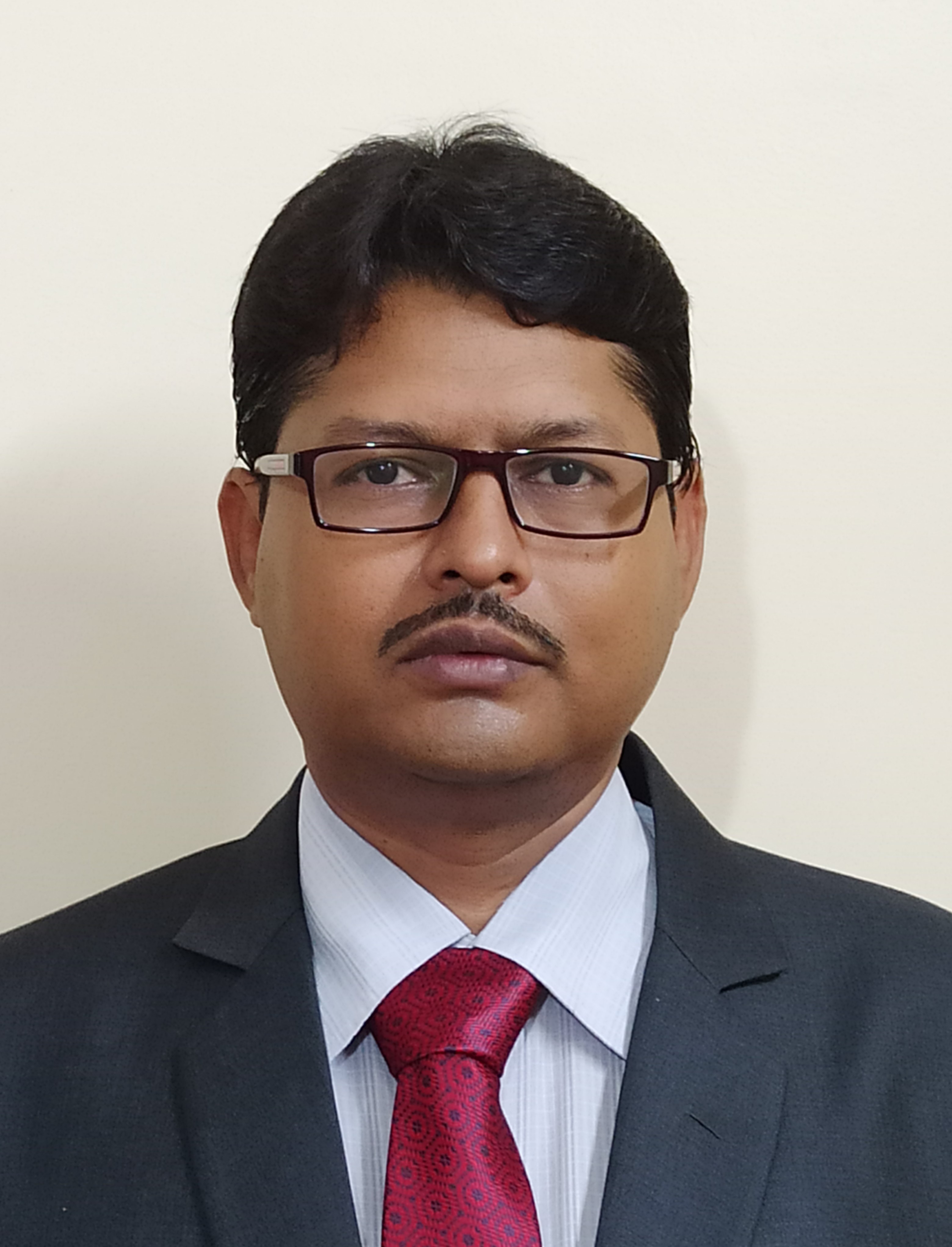}}]{Somnath Dey} is currently working as an Associate Professor and Head in the Department of Computer Science \& Engineering at the Indian Institute of Technology Indore (IIT Indore). He received his B. Tech. degree in Information Technology from the University of Kalyani in 2004. He completed his M.S. (by research) and Ph.D. degree in Information Technology from the School of Information Technology, Indian Institute of Technology Kharagpur, in 2008 and 2013, respectively. His research interest includes biometric security, biometric template protection, biometric cryptosystem and traffic sign detection.
\end{IEEEbiography}
\end{document}